\documentclass[conference]{IEEEtran}
\IEEEoverridecommandlockouts
\usepackage{amsmath,amssymb,amsfonts}
\usepackage{algorithmic}
\usepackage{graphicx}
\usepackage{textcomp}
\usepackage{xcolor}
\usepackage{hyperref}
\usepackage[style=ieee]{biblatex}

\def\BibTeX{{\small\rm B\kern-.05em{\sc i\kern-.025em b}\kern-.08em
    T\kern-.1667em\lower.7ex\hbox{E}\kern-.125emX}}
\begin{document}

\title{PLoc: A New Evaluation Criterion Based on Physical Location for Autonomous Driving Datasets\\

}

\author{\IEEEauthorblockN{Ruining Yang}
\IEEEauthorblockA{\textit{Khoury College of Computer Science} \\
\textit{Northeastern University}\\
Boston, United States \\
yang.ruini@northeastern.edu}
\and
\IEEEauthorblockN{Yuqi Peng}
\IEEEauthorblockA{\textit{Khoury College of Computer Science} \\
\textit{Northeastern University}\\
Boston, United States \\
peng.yuq@northeastern.edu}
}

\maketitle

\begin{abstract}
Autonomous driving has garnered significant attention as a key research area within artificial intelligence. In the context of autonomous driving scenarios, the varying physical locations of objects correspond to different levels of danger. However, conventional evaluation criteria for automatic driving object detection often overlook the crucial aspect of an object's physical location, leading to evaluation results that may not accurately reflect the genuine threat posed by the object to the autonomous driving vehicle. To enhance the safety of autonomous driving, this paper introduces a novel evaluation criterion based on physical location information, termed PLoc. This criterion transcends the limitations of traditional criteria by acknowledging that the physical location of pedestrians in autonomous driving scenarios can provide valuable safety-related information. Furthermore, this paper presents a newly re-annotated dataset (ApolloScape-R) derived from ApolloScape. ApolloScape-R involves the relabeling of pedestrians based on the significance of their physical location. The dataset is utilized to assess the performance of various object detection models under the proposed PLoc criterion. Experimental results demonstrate that the average accuracy of all object detection models in identifying a person situated in the travel lane of an autonomous vehicle is lower than that for a person on a sidewalk. The dataset is publicly available at \href{https://github.com/lnyrlyed/ApolloScape-R.git}{https://github.com/lnyrlyed/ApolloScape-R.git}.


\end{abstract}
\section{Introduction}

Autonomous driving, a prominent research focus in artificial intelligence, has garnered considerable attention from both academia and industry~\cite{nitsch2021out, tian2018lane}. Its impact on future transportation methods~\cite{pakusch2018unintended} within cities is noteworthy, given its capability to execute driving tasks without human intervention. A prerequisite for autonomous driving is highly accurate perception technology~\cite{steinbaeck2017next, marti2019review}, where cameras emerge as pivotal sensors. Each vehicle employs cameras to comprehend the scene and achieve a comprehensive perception of the environment~\cite{panev2018road}. Autonomous driving cars typically capture high-quality images under diverse conditions—varying lighting~\cite{masuda2018rule}, weather~\cite{lee2018development}, and road environments~\cite{uccar2017object,madani2018traffic,geng2017scenario} enabling them to detect objects and signs crucial for safe driving decisions.

Object detection stands out as a primary task in autonomous driving~\cite{chen2017turn,chen2017moving,chen2019surrounding}. In this context, autonomous driving cars must sense their surroundings, classify, and localize objects to make informed driving decisions. While motion planning 

\begin{figure}[h!]
\centering 
\includegraphics[width=0.5\textwidth]{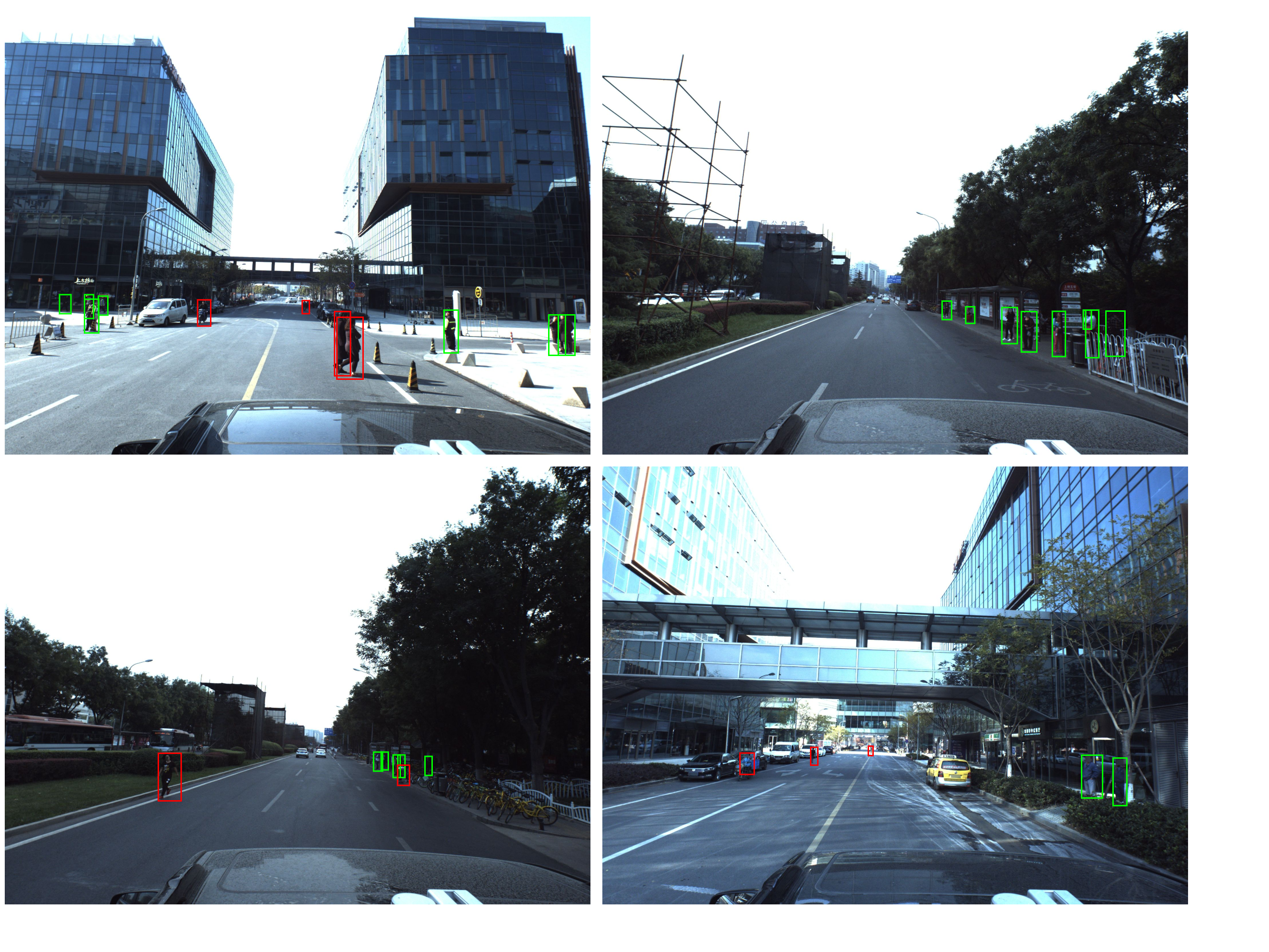} 
\caption{Four images are from the ApolloScape-R dataset. The red bounding box marked people on the road, who are in the path of the driving car, a potential danger position and need to be given more weight. The green bounding box marked people on the sidewalk, who are in a safe position and pose almost no threat to the driving car.}
\label{fig:mainpng}
\end{figure}

 \noindent algorithms can handle scenarios with few obstacles, such as an empty road~\cite{li2013sensor}, in most cases, vehicles need to estimate the trajectories of pedestrians, bicycles, and other vehicles to navigate effectively. Failure to accurately recognize surrounding objects makes it challenging for autonomous driving cars to predict the direction of vehicle and pedestrian movements, leading to potential dangers such as collisions, failure to maintain a safe distance, or accidents with other vehicles.


Object detection algorithms fall into two main categories. The first type is a two-stage detection based on candidate region classification, involving candidate region extraction and subsequent classification with corrected positional coordinates. Notable algorithms in this category include Region-based Convolutional Neural Network (R-CNN)~\cite{girshick2014rich} and its evolved versions Fast R-CNN~\cite{girshick2015fast} and Faster R-CNN~\cite{ren2015faster}, along with several improved iterations~\cite{sun2018face,cheng2018revisiting,dhakal2023sniffer}.

The second category comprises regression-based single-stage object detection algorithms, such as the SSD~\cite{liu2016ssd} and YOLO~\cite{redmon2016you} families. These algorithms are generally faster, predicting object location and class in a single computation, eliminating the need for step-by-step processing. Various improved versions of YOLO~\cite{redmon2018yolov3,bochkovskiy2020yolov4,wang2023yolov7} are the preferred choice for autonomous driving due to its high speed. The advantage of the YOLO algorithm is that it directly predicts the objects in the input image and detects them quickly, which is crucial for the requirements of autonomous driving.

The accuracy of an autonomous driving object detection model hinges on the quality of the dataset used for training. Labeled datasets, focusing on data diversity~\cite{geiger2012we,huang2018apolloscape,yu2020bdd100k,han2021soda10m}, encompassing various cities, roads, weather conditions, and times, are commonly employed. However, diversification alone is insufficient; the evaluation criteria for data validity are paramount. Current datasets often overlook the varying importance of objects at different physical locations in autonomous driving scenarios, treating all objects with equal weight. To better align autonomous driving models with real-world applications, we propose a new evaluation criterion (PLoc) for autonomous driving object detection based on physical location, acknowledging that an individual's importance varies with its location.

Pedestrians, among the most crucial objects in autonomous driving scenarios, hold different importance based on their physical locations. Fig.~\ref{fig:mainpng} provides examples, illustrating that a person on a sidewalk is in a relatively safe position, while a person in a traffic lane is in a highly dangerous area. To implement PLoc, we introduce a re-annotated dataset (ApolloScape-R) based on ApolloScape~\cite{huang2018apolloscape}. This dataset categorizes pedestrians into on-road and not-on-road categories. Subsequently, ten different object detection models are employed for testing. Under PLoc, the average accuracy of all tested models in detecting pedestrians on the road is consistently lower than that for pedestrians off the road. Experimental results highlight variations in the models' ability to detect pedestrians, underscoring the necessity of an evaluation criterion based on physical location.
\section{Related Work}
\subsection{Object Detection}
The object detection task in autonomous driving is mainly to classify and localize objects around the vehicle in real time. With the development of deep learning in recent years, object detection algorithms have evolved from two-stage detection to one-stage detection.

\textbf{Two-stage Object Detection} 
~\cite{girshick2014rich} first proposed R-CNN as the research basis for two-stage target detection algorithms in 2014. R-CNN divides the target detection task into two steps of candidate region extraction and detection, thus increasing the target detection accuracy significantly. However, the fixed input size limitation and a large number of repetitive convolutional operations lead to long training time and slow inference speed of R-CNN. In order to improve the detection speed ~\cite{girshick2015fast} proposed Fast R-CNN in 2015, which is a fast version of R-CNN. Unlike R-CNN, Fast R-CNN introduces region of interest pooling operation to extract candidate features, which significantly improves the detection accuracy, but the process of generating candidate regions greatly inhibits the detection speed of the algorithm. However, the candidate region generation process greatly inhibits the detection speed of the algorithm. To address this problem, ~\cite{ren2015faster} introduces region proposal network in Faster R-CNN, so that region proposal, feature extraction and target detection can be done in a unified network, which greatly improves the detection efficiency.

\textbf{One-stage Object Detection} 
The two-stage type algorithms of RCNN series still have the drawbacks of slow inference speed and complex model, which leads to insufficient detection efficiency and cannot be applied to automatic driving scenarios that require fast detection. In order to solve this problem, ~\cite{redmon2016you} proposed the single-stage target detection algorithm named YOLO in 2015, which is different from the two-stage detection method, and eliminates the region proposal step, and directly performs feature extraction and target detection on the input image, thus predicting the bounding box and the class probability of the target at one time. Then researchers proposed several more efficient versions of YOLO algorithm~\cite{redmon2018yolov3,bochkovskiy2020yolov4,wang2023yolov7} in terms of inference speed and resource consumption to achieve higher accuracy and faster target detection.

\subsection{Autonomous Driving Datasets}
Autonomous driving vehicles rely on a variety of sensors to perceive surrounding objects and make driving decisions. In order for a self-autonomous driving car to make accurate decisions, the model need to be trained on a large amount of effective data, so that the corresponding datasets are created.

\textbf{KITTI}~\cite{geiger2012we}
Object Detection Evaluation 2012 dataset contains real image data captured from urban, rural, and highway scenes, with up to 15 vehicles and 30 pedestrians in each image. It consists of 7481 training images and 7518 test images. All images are in color and saved as png.

\textbf{Apollo}~\cite{huang2018apolloscape}
presented by Baidu in 2019 consists of 73 city street view videos from all over China and under different weather conditions. It contains over 140K images and their corresponding 2D/3D semantic annotations. Appllo dataset has 26 categories including cars, bicycles, pedestrians, buildings, streetlights and so on.

\textbf{BDD100K}~\cite{yu2020bdd100k}
launched in 2020 by University of California, Berkeley, contains 100k videos and 10 tasks. The videos are annotated with a variety of annotations including image tagging, lane detection, drivable area segmentation, road object detection, semantic segmentation, instance segmentation, multi-object detection tracking, multi-object segmentation tracking, domain adaptation, and imitation learning. BDD100K has geographic, environmental, and weather diversity and ten categories of ground truth labels.

\textbf{SODA10M}~\cite{han2021soda10m}
launched in 2021 by Huawei Lab, which includes 10 million unlabeled images of road scenes collected from 32 cities, and 20k images with complete annotation, annotated with six major types of human and vehicle scene categories. The unlabeled images were selected from 32 different cities, covering most regions of China. Additionally, the images encompass a variety of road scenes, weather conditions and times of the day.

\section{New Evaluation criterion Based on Physical Location}
Existing object detection datasets and models fall short of meeting the demands of real-world autonomous driving scenarios due to their oversight of the physical location information of objects. In this section, we address the limitations of existing evaluation criteria in \S\ref {subsec:limitation} and propose a new evaluation criteria(PLoc) based on physical location \S\ref{subsec:criteria}. PLoc focuses on improving the ability of the autonomous driving model to recognize objects at different locations, in order to improve safety. In addition, a new ApolloScape re-annotated dataset ApolloScape-R \S\ref{subsec:dataset} is proposed to implement PLoc and validate its effectiveness with different object detection models.
\subsection{Limitations of Existing Evaluation Criteria}
\label{subsec:limitation}
A highly effective deep learning object detection model hinges not only on algorithmic design but also on the quality of training data, directly influencing the model's performance. Consequently, it is crucial to focus on enhancing the training data's utility to improve model accuracy. In autonomous driving scenarios, the accuracy of object detection models is particularly critical as it directly impacts the ability of autonomous vehicles to make safe driving decisions [15].  Inaccurate identification of pedestrians by these models may result in vehicles failing to avoid them promptly, potentially leading to traffic accidents.

However, prevailing object detection datasets typically treat all pedestrians equally, disregarding the varying importance based on their physical positions. In the real world, the position of pedestrians directly influences the decision-making process of autonomous vehicles. For instance, pedestrians on the roadway pose a more immediate threat to autonomous vehicles compared to those on sidewalks. Neglecting this distinction, models fail to recognize the importance of pedestrians in different locations, diminishing the efficiency of driving decisions.

\begin{figure}[h!]
\centering 
\includegraphics[width=0.5\textwidth]{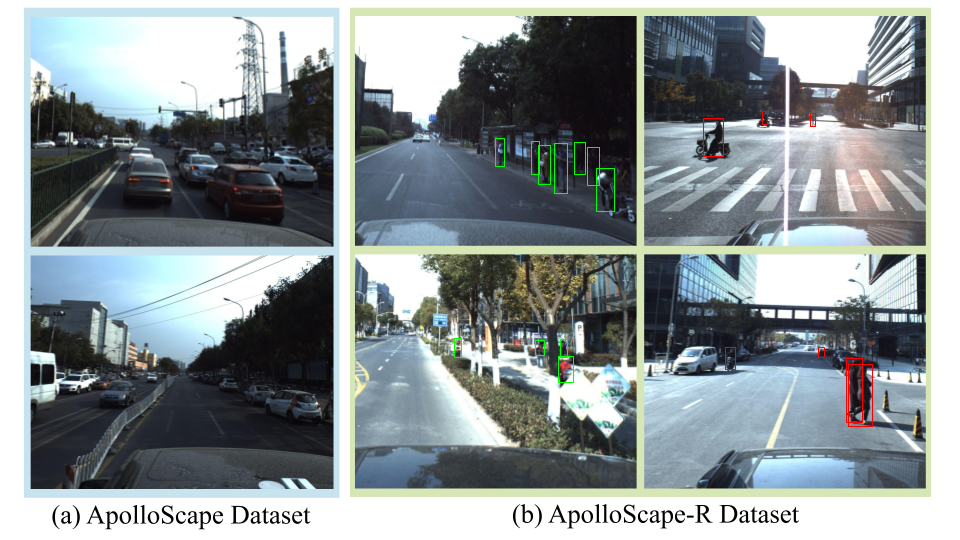} 
\caption{The ApolloScape~\cite{huang2018apolloscape} dataset has some images that contain only vehicles, the two images on the left are examples from the ApolloScape~\cite{huang2018apolloscape} dataset, whereas we focus more on pedestrians as objects. Therefore, the images containing people were selected and re-annotated to form the new ApolloScape-R dataset. People in different physical locations are separated into two categories. Two images in the middle is labeled with a green bounding box. Pedestrians are on the sidewalk and pose little threat to the autonomous driving cars. On the right images, people who labeled with a red bounding box are in the direction of autonomous driving cars and are more likely to be involved in an accident.}
\label{fig:dataset}
\end{figure}

\subsection{Proposal of New Evaluation Criteria}
\label{subsec:criteria}
Because of the limitations of existing measurements standard, we propose a new evaluation criteria(PLoc) that combines physical location of pedestrians into the consideration. It is beneficial for adapting models more effectively to real-world application scenarios by changing datasets. In PLoc, pedestrians are categorized into two different classes based on their physical location in the autonomous driving scenario: 
\textbf{1. Pedestrians on the roadway: } 
Since these pedestrians are in the direct path of the autonomous driving car, there is a high possibility of accidents occurring and get pedestrians in danger. \textbf{2. Pedestrians on the sidewalk: }
Although overall driving environment awareness is still important, these pedestrians pose less threat to autonomous driving cars. Therefore, these types of pedestrians are given a lower weight under the guidance of PLoc. This can be achieved by setting different weight for pedestrian, such as a weight of 1.0 for pedestrians on the roadway and 0.5 for pedestrians on the sidewalk. 

Generally, the loss function of an object detection model comprises three main components: Box Loss, Classification Loss, and Distribution Focal Loss. These components are utilized to assess the accuracy of the bounding box, the effectiveness of target identification, and the ability to handle category imbalance and indistinguishable samples. The mathematical expression of this loss function is as follows:
\begin{equation}
\mathcal{L}(\theta)=\frac{1}{N} \left( \lambda_{\text {box}} \mathcal{L}_{\text{box}}(\theta)+\lambda_{\text{cls}} \mathcal{L}_{\text{cls}}(\theta) + \lambda_{\text{dfl}} \mathcal{L}_{\text{dfl}}(\theta) \right)
\label{eq:loss}
\end{equation}
where:
\begin{itemize}
\item $N$ is the total number of cells.
\item $\lambda_{\text{box}}$, $\lambda_{\text{cls}}$ and $\lambda_{\text{dfl}}$ are the weights of Box Loss, Classification Loss and Distribution Focal Loss respectively.
\item $\mathcal{L}_{\text{box}}(\theta)$, $\mathcal{L}_{\text{cls}}(\theta)$ and $\mathcal{L}_{\text{dfl}}(\theta)$ represent the functions of Box Loss, Classification Loss and Distribution Focal Loss respectively. 
\end{itemize}

Building on the concept we just proposed, we categorize the pedestrians in the dataset into two groups: those on the road and those on the sidewalks. We use Eq. \ref{eq:loss} to calculate the loss for each type of pedestrian separately. The total loss function of the new model can then be represented as a weighted sum of these two losses. Specifically, the loss attributed to pedestrians on the road is expressed as follows:

\begin{equation}
\mathcal{L_\text{r}}(\theta)=\frac{1}{N_\text{r}} \left( \lambda_{\text {box}} \mathcal{L}_{\text{rbox}}(\theta)+\lambda_{\text{cls}} \mathcal{L}_{\text{rcls}}(\theta) + \lambda_{\text{dfl}} \mathcal{L}_{\text{rdfl}}(\theta) \right) 
\end{equation}
The loss caused by pedestrians on the sidewalks is partially expressed as:
\begin{equation}
\mathcal{L_\text{s}}(\theta)=\frac{1}{N_\text{s}} \left( \lambda_{\text {box}} \mathcal{L}_{\text{sbox}}(\theta)+\lambda_{\text{cls}} \mathcal{L}_{\text{scls}}(\theta) + \lambda_{\text{dfl}} \mathcal{L}_{\text{sdfl}}(\theta) \right) 
\end{equation}
Then, the total loss function is in the form of:
\begin{equation}
\mathcal{L}(\theta)=\lambda_{\text{r}}\mathcal{L_\text{r}}(\theta) + \lambda_{\text{s}}\mathcal{L_\text{s}}(\theta)
\end{equation}
where $\lambda_{\text{r}}$ and $\lambda_{\text{s}}$ are the weights of the loss caused by pedestrians on the road and sidewalks respectively. 

By adjusting the weights of the two losses, the model can focus more on the loss caused by certain pedestrians. In the previous example, if we give a higher weight (like 1) to pedestrians on the road and a lower weight (like 0.5) to those on the sidewalk, then the model will focus more on issues caused by pedestrians on the road. By making this distinction, autonomous driving models can be motivated to pay more attention on pedestrians in the roadway, while still maintaining the ability to recognize all pedestrians.

\subsection{ApolloScape-R Dataset}
\label{subsec:dataset}
To implement PLoc and validate its applicability in real-world scenarios, we introduce the ApolloScape-R dataset, derived from the ApolloScape [28] dataset. Specifically, we selected images from ApolloScape containing pedestrians and re-annotated them based on PLoc, considering the physical location of pedestrians. Subsequently, we conducted experiments on ApolloScape-R using ten object detection models to assess their effectiveness in detecting pedestrians at different locations under PLoc. Examples of ApolloScape-R are depicted in Fig.~\ref{fig:dataset}.

\section{Experimental and Discussion}

\begin{table}[h!]
\centering
\caption{Person on Road Detection Results}
\begin{tabular}{l|l|rrr}
\hline
       Model &  Publication &   mAP &  $mAP_{50}$ &  $mAP_l$ \\
\hline
Faster Rcnn~\cite{ren2015faster} & NeurIPS 2015 & 0.045 &   0.085 &   0.137 \\
   RetinaNet~\cite{lin2017focal} &    ICCV 2017 & 0.041 &   0.078 &   0.169 \\
Cascade Rcnn~\cite{cai2019cascade} &    CVPR 2018 & 0.045 &   0.080 &   0.148 \\
        FCOS~\cite{tian2019fcos} &    ICCV 2019 & 0.053 &   0.099 &   0.168 \\
   RepPoints~\cite{yang2019reppoints} &    ICCV 2019 & 0.044 &   0.085 &   0.155 \\
        ATSS~\cite{zhang2020bridging} &    CVPR 2020 & 0.047 &   0.082 &   0.161 \\
Deformable-DETR~\cite{carion2020end} &    ECCV 2020 & 0.030 &   0.064 &   0.107 \\
 Sparse Rcnn~\cite{sun2021sparse} &    CVPR 2021 & 0.037 &   0.078 &   0.107 \\
      YOLOX~\cite{ge2021yolox} &    CVPR 2021 & 0.061 &   0.107 &   0.184 \\
         DDQ~\cite{zhang2023dense} &    CVPR 2023 & 0.066 &   0.125 &   0.164 \\
\hline
\multicolumn{5}{l}{All the models are trained with the ResNet-50~\cite{he2016deep} as the backbone,}\\
\multicolumn{5}{l}{with the exception of YOLOX~\cite{ge2021yolox}, which trained with DarkNet53~\cite{redmon2018yolov3}.}
\end{tabular}
\label{table:on_road}
\end{table}

\begin{table}[h!]
\centering
\caption{Person Not on Road Detection Results}
\begin{tabular}{l|l|rrr}
\hline
       Model &  Publication &   mAP &  $mAP_{50}$ &  $mAP_l$ \\
\hline
 Faster Rcnn~\cite{ren2015faster} & NeurIPS 2015 & 0.061 &   0.137 &   0.205 \\
   RetinaNet~\cite{lin2017focal} &    ICCV 2017 & 0.043 &   0.102 &   0.190 \\
Cascade Rcnn~\cite{cai2019cascade} &    CVPR 2018 & 0.072 &   0.155 &   0.240 \\
        FCOS~\cite{tian2019fcos} &    ICCV 2019 & 0.069 &   0.154 &   0.221 \\
   Reppoints~\cite{yang2019reppoints} &    ICCV 2019 & 0.061 &   0.138 &   0.224 \\
        ATSS~\cite{zhang2020bridging} &    CVPR 2020 & 0.060 &   0.131 &   0.238 \\
        Deformable-DETR~\cite{carion2020end} &    ECCV 2020 & 0.031 &   0.098 &   0.119 \\
 Sparse Rcnn~\cite{sun2021sparse} &    CVPR 2021 & 0.060 &   0.138 &   0.184 \\
      YOLOX~\cite{ge2021yolox} &    CVPR 2021 & 0.080 &   0.164 &   0.225 \\
         DDQ~\cite{zhang2023dense} &    CVPR 2023 & 0.120 &   0.236 &   0.262 \\
\hline
\multicolumn{5}{l}{All the models are trained with the ResNet-50~\cite{he2016deep} as the backbone,}\\
\multicolumn{5}{l}{with the exception of YOLOX~\cite{ge2021yolox}, which trained with DarkNet53~\cite{redmon2018yolov3}.}
\end{tabular}
\label{table:not_on_road}
\end{table}

\subsection{Experimental Result}
To validate the new evaluation criteria PLoc, we proposed a new dataset that includes the physical location information of pedestrians and conducted experiments using 10 popular object detection models. Experimental result shows in Table~\ref{table:on_road},~\ref{table:not_on_road}. These models include: ATSS~\cite{zhang2020bridging}, Cascade R-CNN~\cite{cai2019cascade}, DDQ~\cite{zhang2023dense}, Deformable-DETR~\cite{carion2020end}, Faster R-CNN~\cite{ren2015faster}, FCOS~\cite{tian2019fcos}, RepPoints~\cite{yang2019reppoints}, RetinaNet~\cite{lin2017focal}, Sparse R-CNN~\cite{sun2021sparse}, and YOLOX~\cite{ge2021yolox}. We categorized pedestrian targets into two types, including pedestrians on the road and pedestrians not on the road, in order to get the performance of different models in recognizing these two types of pedestrians. In the experiments, we used the standard mean Average Precision (mAP) and calculated at different Intersection over Union (IoU) thresholds, such as mAP, $\text{mAP}_{\text{50}}$ and $\text{mAP}_{\text{l}}$ (for large-sized objects) as evaluation metrics. By comparing three different mAP values, all tested models in the "Pedestrians on the Road" category have better performance in the "Pedestrians not on the Road".

Among them, the DDQ~\cite{zhang2023dense} model has the best pedestrian detection performance, be cause its three mAP values for person not on road are higher than other models, which are 0.120, 0.236 and 0.262 respectively. However, when this model detects person on the road, its detection effect drops significantly, only 0.066, 0.125 and 0.164. This means that the model has obvious performance advantages in detecting pedestrians not on the road compared to detecting pedestrians on the road. Similar significant performance differences also appear between the Cascade-RCNN~\cite{cai2019cascade} model and Sparse-RCNN~\cite{sun2021sparse} model. Although compared with this huge difference, the performance difference of ASTT~\cite{zhang2020bridging}, Faster-RCNN~\cite{ren2015faster}, FCOS~\cite{tian2019fcos}, RepPoints~\cite{yang2019reppoints} and YOLOX~\cite{ge2021yolox} models in detecting two different pedestrians has been reduced, but they are still better at detecting pedestrians on the road. The Deformable-DETR~\cite{carion2020end} and RetinaNet~\cite{lin2017focal} models have shown relatively good performance consistency in the detection of different types of pedestrians. They have similar effects in detecting pedestrians on the road and pedestrians not on road. However, their mAP values are at a low level compared to other models, which means that although their performance does not fluctuate significantly due to changes in pedestrian locations, their detection effects are not outstanding.

\begin{figure}[t]
\centering 
\includegraphics[width=0.5\textwidth]{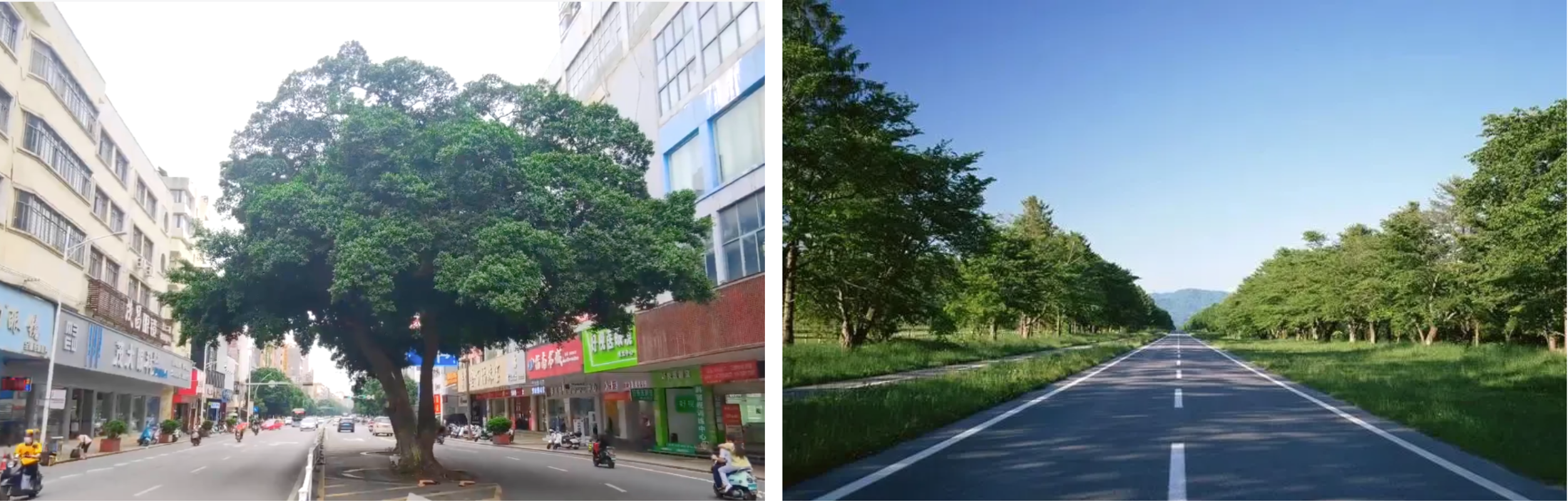} 
\caption{The tree on the left is in the middle of the vehicle's driving path and can directly affect driving safety. Autonomous driving vehicles need to accurately recognize trees to avoid making mistakes that could lead to danger. The tree on the right appears only as a background and does not pose a threat to the autonomous driving vehicle's decisions.}
\label{fig:tree}
\end{figure}


\subsection{Discussion}
Experiments across different models confirm a significant difference in the accuracy of object detection models for pedestrians at different physical locations. This observation extends beyond pedestrian detection, constituting a new evaluation criterion for autonomous driving object detection based on physical location. This location-based evaluation measurement can be further extended to other object detection tasks in the realm of autonomous driving, enhancing the accuracy of these models in real-world applications.

For instance, trees in different locations may impact a vehicle's movement differently, as illustrated in Fig. ~\ref{fig:tree}. In most scenarios, the tree on the roadside may only be part of the environment and does not directly affect the driving path. While, in some special cases, a tree may grow in the middle of the road, which requires the detection model to be able to accurately recognize and avoid it. Therefore, evaluating the importance of trees depending on their physical location can help autonomous driving vehicles make better decisions.

Autonomous driving vehicles are not only influenced by objects located on the road, but signal-type objects also play a crucial role in driving decisions. Distinguishing between different signals based on their location is essential to ensure that autonomous driving cars follow traffic rules correctly. For instance, when a vehicular traffic signal is in a position similar to a pedestrian crossing signal, the vehicular traffic signal should be given more attention to ensure accurate signal recognition.

Additionally, considering the design of entrances and exits on a road, traffic signs in different physical locations can pose a threat to the accurate decision-making of an autonomous driving car. The autonomous driving object detection model should accurately recognize signs at different locations to avoid mistakes that could lead to dangerous situations.

In summary, the physical location-based evaluation criterion holds practical application value, effectively improving the accuracy of the autonomous driving object detection model and enhancing overall safety. Moreover, this criterion can be generalized to various fields such as urban planning, intelligent monitoring, and robot navigation, providing new perspectives and methods for solving practical problems in complex environments.
\section{Conclusion}
This research addresses one of the primary challenges in autonomous driving: the accurate detection of objects that pose a threat to autonomous driving vehicles. We introduce a new evaluation criterion based on the physical location for autonomous driving datasets, aiming to overcome the limitations of existing criteria. Using pedestrians as a case study, this paper presents the ApolloScape-R dataset, which categorizes pedestrians based on different physical locations. We evaluate ten popular object detection models on this new criterion. The experimental results demonstrate the effectiveness of the proposed evaluation criterion, with all tested models performing better in detecting pedestrians not on the road compared to those on the road. Furthermore, we propose the extension of this evaluation criterion to different objects and signaling signs in autonomous driving scenarios. This extension has the potential to enhance the practicality and safety of autonomous driving technologies. By considering the physical location of objects, such as trees, signals, and road signs, autonomous driving models can make more informed and accurate decisions, contributing to the overall improvement of autonomous driving systems in real-world applications. This research opens avenues for further exploration and development in the field, with the ultimate goal of making autonomous driving safer and more reliable.

\printbibliography[title={References}]

\end{document}